\documentclass[10pt,twocolumn,letterpaper]{article}

\usepackage{cvpr}              
\usepackage[accsupp]{axessibility} 

\usepackage{graphicx}
\usepackage{amsmath}
\usepackage[utf8]{inputenc}
\usepackage{amssymb}
\usepackage{booktabs}
\usepackage{makecell}
\usepackage{lipsum}
\usepackage{multirow}
\usepackage{multicol}
\usepackage{adjustbox}
\makeatletter
\newcommand\fs@norules{\def\@fs@cfont{\bfseries}\let\@fs@capt\floatc@ruled
  \def\@fs@pre{}%
  \def\@fs@post{}%
  \def\@fs@mid{\kern3pt}%
  \let\@fs@iftopcapt\iftrue}
\makeatother

\usepackage[pagebackref,breaklinks,colorlinks]{hyperref}

\usepackage[capitalize]{cleveref}
\crefname{section}{Sec.}{Secs.}
\Crefname{section}{Section}{Sections}
\Crefname{table}{Table}{Tables}
\crefname{table}{Tab.}{Tabs.}

\def\cvprPaperID{14525} 
\def\confName{CVPR}
\def\confYear{2024}

\begin{document}

\title{RadarDistill: Boosting Radar-based Object Detection Performance via Knowledge Distillation from LiDAR Features}

\author{{Geonho Bang$^{1}$\thanks{Equal contributions}
\qquad
Kwangjin Choi$^{1}$\footnotemark[1]
\qquad
Jisong Kim$^{1}$
\qquad
Dongsuk Kum$^{2}$
\qquad
Jun Won Choi$^{3}$\thanks{Corresponding author}} \\
$^{1}$Hanyang University, Korea \qquad $^{2}$KAIST, Korea \qquad $^{3}$Seoul National University, Korea
\and
{\tt\small $^{1}$\{ghbang, kjchoi, jskim\}@spa.hanyang.ac.kr} \and {\tt\small $^{2}$dskum@kaist.ac.kr} \and {\tt\small $^{3}$junwchoi@snu.ac.kr}
}

\maketitle


\begin{abstract}
The inherent noisy and sparse characteristics of radar data pose challenges in finding effective representations for 3D object detection. In this paper, we propose RadarDistill, a novel knowledge distillation (KD) method, which can improve the representation of radar data by leveraging LiDAR data. 
RadarDistill successfully transfers desirable characteristics of LiDAR features into radar features using three key components: Cross-Modality Alignment (CMA), Activation-based Feature Distillation (AFD), and Proposal-based Feature Distillation (PFD). 
CMA enhances the density of radar features by employing multiple layers of dilation operations, effectively addressing the challenge of inefficient knowledge transfer from LiDAR to radar. AFD selectively transfers knowledge based on regions of the LiDAR features, with a specific focus on areas where activation intensity exceeds a predefined threshold.
PFD similarly guides the radar network to selectively mimic features from the LiDAR network within the object proposals.
Our comparative analyses conducted on the nuScenes datasets demonstrate that RadarDistill achieves state-of-the-art (SOTA) performance for radar-only object detection task, recording \textbf{20.5\%} in mAP and \textbf{43.7\%} in NDS. Also, RadarDistill significantly improves the performance of the camera-radar fusion model.
\end{abstract}

\begin{figure}[t]
\centering  
\includegraphics[width=7.2cm]{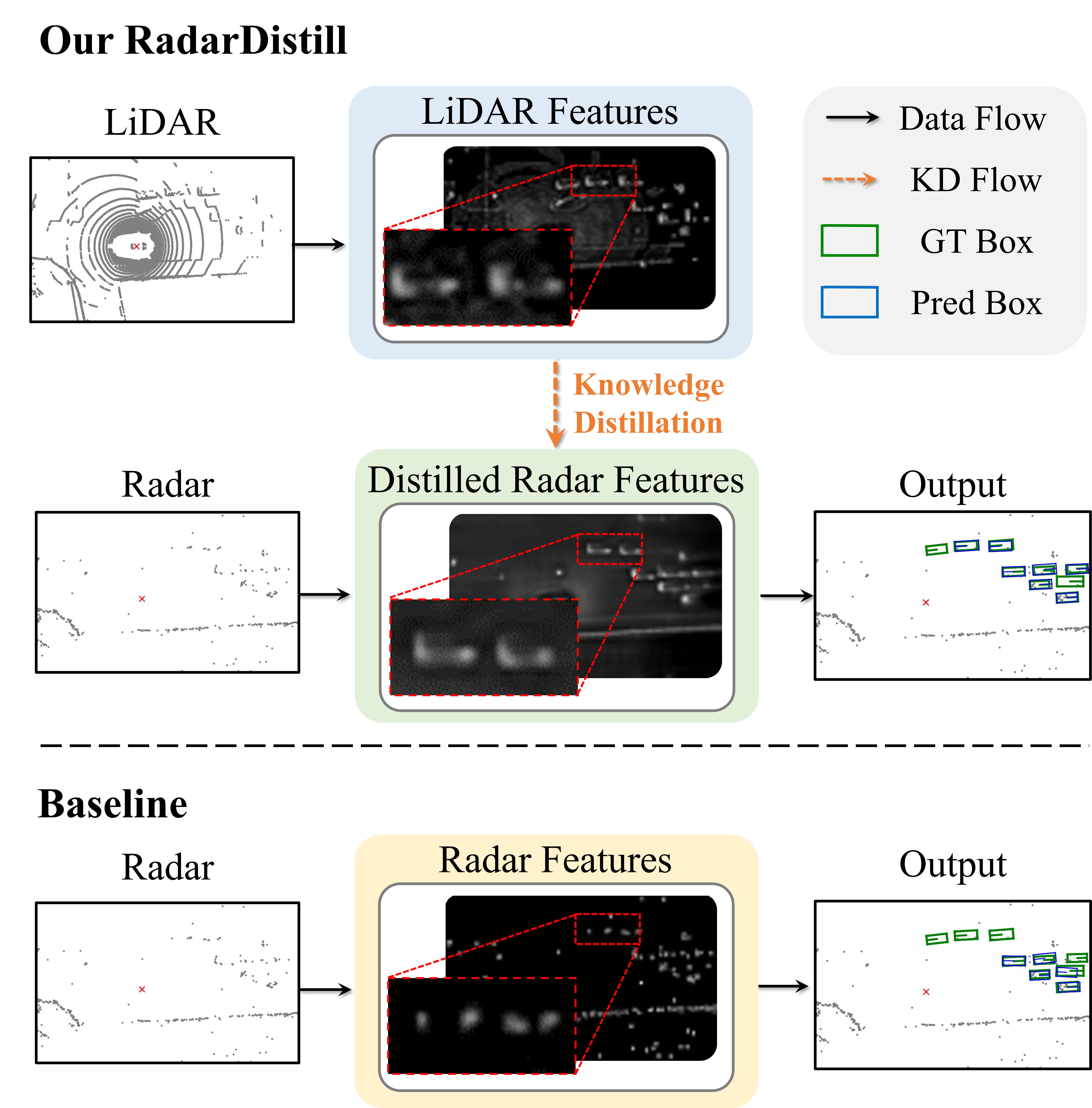}
\caption{\textbf{Illustration of Proposed RadarDistill.} Our RadarDistill method facilitates knowledge transfer from LiDAR features to radar features, enhancing the quality of radar features for Bird's Eye View (BEV) object detection.}
\label{keyconcept}  
\end{figure}

\section{Introduction}
\label{sec:intro}
While 3D perception based on camera and LiDAR sensors has been widely studied, radar sensors are now gaining attention due to their affordability and reliability in adverse weather conditions. 
Radar sensors can locate objects in a Bird's Eye View (BEV) and also measure their radial velocity using Doppler frequency analysis. However, when compared to LiDAR or camera sensors, radar's major limitations are its lower spatial resolution and a higher likelihood of false positives due to multi-path reflections. For decades, traditional object detection and tracking methods, based on manually crafted models, have been developed to overcome these limitations by many radar manufacturers. While deep neural networks (DNNs) have considerably improved 3D perception in camera and LiDAR sensors, similar advancements have not been mirrored in radar sensor-specific architectures.
There are only a handful of studies that have applied deep neural networks to radar data. For instance, KPConvPillars \cite{kpconvpillar} and Radar-PointGNN \cite{radar-pointgnn} leveraged KPConv \cite{kpconv} and graph neural networks, respectively, to detect objects using radar point clouds. However, these methods have not yet achieved the level of significant improvements realized with camera or LiDAR data. Recently, it was shown that radar can be effectively fused with camera or LiDAR data to enhance the robustness of 3D object detection \cite{craft, crn, rcbev, centerfusion, bi-lrfusion, rcm-fusion}.

This paper focuses on improving the performance of radar-based 3D object detection using deep neural networks. We note that the limited performance of radar is largely due to the challenges in finding effective representations, given the sparse and noisy nature of radar measurements. Inspired by the remarkable success of deep models that encode LiDAR point clouds, our goal is to transfer the knowledge extracted from a LiDAR-based model to enhance a radar-based model.

Recently, knowledge distillation (KD) techniques have shown success in transferring knowledge from one sensor modality to another, thereby improving the representation quality of the target model. 
To date, various KD methods have been introduced in the literature \cite{ligastereo, monodistill, cmkd, bevdistill, unidistill, x3kd, distillbev, s2m2ssd}.
BEVDistill \cite{bevdistill} transforms both LiDAR and cameras features into a BEV format, enabling the transfer of spatial knowledge from LiDAR features to camera features. DistillBEV \cite{distillbev} utilizes the prediction results from either LiDAR or LiDAR-camera fusion models to distinguish between foreground and background, guiding the student model to focus on KD in essential areas. S2M2-SSD \cite{s2m2ssd} determines key areas based on the student model's predictions and transfers information obtained from a LiDAR-camera fusion model in the key areas. Apart from these approaches, UniDistill \cite{unidistill} employs a universal cross-modality framework that enables knowledge transfer among diverse modalities. This framework is adaptable to different modality pairings, including camera-to-LiDAR, LiDAR-to-camera, and (camera+LiDAR)-to-camera settings.

In this paper, we propose RadarDistill, a novel KD framework designed to enhance the representation of radar data, leveraging LiDAR data. 
Our study shows that by employing a radar encoding network as a student network and a LiDAR encoding network as a teacher network, our KD framework effectively produces radar features akin to the dense and semantically rich LiDAR features.
Although both LiDAR data and its encoding network are used to enhance radar features during the training phase, they are not required in the inference phase.

Our proposed RadarDistill is designed based on three main ideas: 1) Cross-Modality Alignment (CMA), 2) Activation-based Feature Distillation (AFD), and 3) Proposal-based Feature Distillation (PFD). 
Our study indicates that transferring knowledge from LiDAR to radar features is difficult due to the inherent sparsity of radar data, which complicates the alignment with the more densely distributed LiDAR features.
To address this problem, CMA boosts the student network's capacity and simultaneously increases the ratio of active radar features by implementing multiple layers of dilation operations.

The proposed AFD and PFD minimize the distribution gap between the intermediate features produced by the radar and LiDAR encoding networks.
Initially, AFD conducts Activation-aware Feature Matching on low-level features. Specifically, it divides both radar and LiDAR features into active and inactive regions according to activation intensity of each features and constructs KD losses for each region separately. By assigning greater importance to the KD loss linked with sparsely distributed active regions, AFD enables the network to concentrate on transferring knowledge for significant features.

Next, PFD implements Proposal-level Selective Feature Matching, aimed at narrowing the differences from features associated with the proposals generated by the radar detection head.
PFD directs the radar network to generate object features that are similar in shape to the high-level LiDAR features for accurately detected proposals.
Conversely, for misdetected proposals such as false positives, the model is guided to suppress falsely activated features, reflecting low activation of the LiDAR features.

Combining all these ideas, our RadarDistill achieves \text{+}{\bf15.6$\%$} gain in mAP and a \text{+}{\bf29.8$\%$} gain in NDS over the current state-of-the-art (SOTA) performance of radar-only object detection methods on nuScenes benchmark \cite{nuscenes}. We also show that when the radar features enhanced by RadarDistill are integrated into a radar-camera fusion model, significant performance improvement is achieved.

The key contributions of our work are as follows:
\begin{itemize}
    \item Our study is the first to demonstrate that radar object detection can be substantially improved using LiDAR data during the training process. Our qualitative results in Fig. \ref{keyconcept} highlight that the radar features acquired through RadarDistill successfully mimic those of LiDAR, leading to enhanced object detection and localization.
    \item Our findings reveal that the CMA is a crucial element of RadarDistill. In its absence, we observed a significant drop in performance enhancement. According to our ablation study, CMA plays a pivotal role in resolving inefficient knowledge transfer caused by the different densities of radar and LiDAR point clouds.
    \item We propose two novel KD methods, AFD and PFD. These methods bridge the discrepancy between radar and LiDAR features, operating at two separate feature levels and utilizing KD losses specifically designed for each level.
    \item RadarDistill achieves the state-of-the-art performance in the radar-only object detector category on the nuScenes benchmark. It also achieves a significant performance boost for camera-radar fusion scenarios.
\end{itemize}


\begin{figure*}[t]
\centering  
\includegraphics[width=15.5cm]{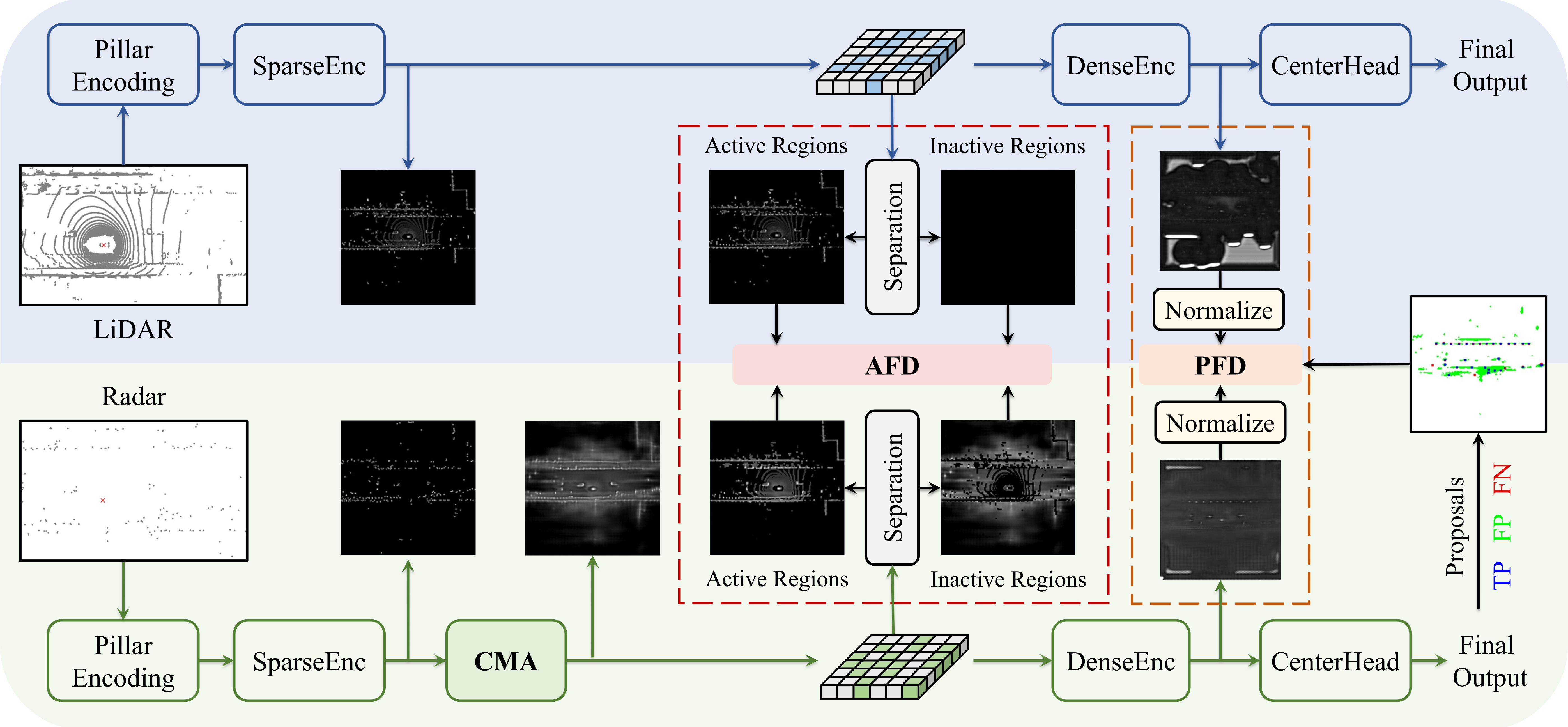}
\caption{\textbf{Overall architecture of RadarDistill.} The input point clouds from each modality are independently processed through Pillar Encoding followed by SparseEnc to extract low-level BEV features. CMA is then employed to densify the low-level BEV features in the radar branch. AFD then identifies active and inactive regions based on both radar and LiDAR features and minimizes their associated distillation losses. Subsequently, PFD conducts knowledge distillation based on proposal-level features obtained from DenseEnc. Note that the LiDAR branch is solely utilized during the training phase to enhance the radar pipeline and is not required during inference.}
\label{architecture}  
\end{figure*}

\section{Related Works}
\label{sec:related work}

\subsection{Radar-based 3D Object Detection} 
Radar-only 3D object detection models have employed backbone models adopted from various LiDAR-based detectors to suit the specific needs of radar data. Radar-PointGNN \cite{radar-pointgnn} utilized GNNs \cite{pointgnn} to effectively extract features from sparse radar point clouds through a graph representation. KPConvPillars \cite{kpconvpillar} introduced a hybrid architecture that combined grid-based and point-based approaches for radar-only 3D object detection. 
Recent studies have concentrated on detection techniques that combine radar with LiDAR or cameras \cite{centerfusion, craft, crn, rcbev, radarnet, bi-lrfusion}. These methods aimed to complement the limitations of each sensor by using the data obtained from radars. 
RadarNet \cite{radarnet} employed a voxel-based early fusion and an attention-based late fusion approach to integrate radar and LiDAR data. 
RCM-Fusion \cite{rcm-fusion} fused radar and camera data at both the feature level and instance level to fully utilize the potential of radar information. CRN \cite{crn} leveraged radar data to transform camera image features into a BEV view and then combined these transformed camera features with the radar BEV features using multi-modal deformable attention.

\subsection{Knowledge Distillation}
Knowledge distillation, as a strategy for model compression, transfers the information from a larger teacher model with greater capacity to a smaller student model. The student model mimics the teacher’s intermediate features \cite{fitnets}, prediction logit \cite{hinton}, or activation boundary \cite{activation} to acquire knowledge from the teacher model. KD approaches have recently been extended from image classification to object detection \cite{chen2017learning, li2017mimicking, fgfi, defeat, fgd, pgd}. Defeat \cite{defeat} utilized KD by decoupling features to foreground and background with ground truth. FGD \cite{fgd} employed spatial and channel attention for 2D object detection, guiding the student model to focus on critical pixels and channels.

In 3D object detection, KD was applied to reduce model complexity in 2D detection \cite{sparsekd, rdd, pointdistill}, or to enhance detection performance through cross-modality knowledge distillation \cite{ligastereo, monodistill, cmkd, uvtr, s2m2ssd, bevdistill, unidistill, x3kd, distillbev}. SparseKD \cite{sparsekd} focused on KD for primary regions identified by the teacher model’s predictions, and the student model was initialized with the weights of the teacher to enhance feature extraction. MonoDistill \cite{monodistill} projected LiDAR point clouds onto the image plane and used a 2D CNN to develop an `image-version' LiDAR-based model, effectively bridging the gap in feature representations between LiDAR and camera. BEVDistill \cite{bevdistill} projected LiDAR and camera features into the BEV space, effectively transferring 3D spatial knowledge through dense feature distillation and sparse instance distillation. UniDistill \cite{unidistill} proposed a universal framework that was suitable for various teacher-student pairs. UniDistill projected both teacher and student features into BEVs and applied KD on the object features within each ground truth box. 

As compared to the existing studies on cross-modality KD methods, our study focuses on developing KD specifically tailored for radar object detection, considering the sparse and noisy nature of radar data.

\section{Method}
\label{sec: methods}
In this section, we present the details of the proposed KD framework, RadarDistll. Fig. \ref{architecture} illustrates the overall architecture of RadarDistill. In Section \ref{subsec: preliminary}, we provide a brief overview of the baseline model employed as both teacher and student models.
In Section \ref{subsec: CMA}, we describe the details of Cross-Modality Alignment, a module for densifying radar features. In Section \ref{subsec: AFD}, we present the AFD, a novel Activation-aware Feature Matching approach. Finally Section \ref{subsec: PFD} outlines the PFD, a Proposal-level Selective Feature Matching approach.

\subsection{Preliminary}
\label{subsec: preliminary}
We use PillarNet \cite{pillarnet} as our baseline model for both LiDAR and radar detectors. PillarNet organizes both radar and LiDAR point clouds using a 2D pillar structure of the same size. It then generates two BEV pillar features $F^{2D}_{ldr}$ and $F^{2D}_{rdr}$, utilizing separate pillar encoding networks in the LiDAR and radar branches, respectively.
We define the low-level BEV features $F^{(l)}_{ldr}$ and $F^{(l)}_{rdr}\in \mathbb{R}^{C \times \frac{H}{8} \times \frac{W}{8}}$ obtained by encoding the pillar features $F^{2D}_{ldr}$ and $F^{2D}_{rdr}$ as
\begin{align}   
 \label{eq:sparseenc}
 & F^{(l)}_{mod} = SparseEnc(F^{2D}_{mod}),
\end{align}
where $SparseEnc(\cdot)$ denotes a {\it 2D sparse convolution-based encoder} (SparseEnc) \cite{spconv} and $mod$ represents the detector modality, i.e., $mod \in \{ldr, rdr \}$. 
Similarly, we describe the high-level BEV features $F^{(h)}_{ldr}$ and $F^{(h)}_{rdr}\in \mathbb{R}^{C \times \frac{H}{8} \times \frac{W}{8}}$ formed by encoding the low-level BEV features $F^{(l)}_{ldr}$ and $F^{(l)}_{rdr}$, respectively as
\begin{align}   
 \label{eq:denseenc}
 & F^{(h)}_{mod} = DenseEnc(F^{(l)}_{mod}),
\end{align}
where $DenseEnc(\cdot)$ denotes a {\it 2D dense convolution-based encoder} (DenseEnc). 
The high-level BEV features are further processed through a CenterHead network \cite{centerpoint} to generate the final prediction heatmaps for classification, regression, and IoU scoring
\begin{align}   
 \label{eq:centerhead}
 & H^{cls}_{mod}, H^{reg}_{mod}, H^{IoU}_{mod} = CenterHead(F^{(h)}_{mod}).
\end{align}
Note that the pipeline in the LiDAR branch serves as the teacher model, while the one in the radar branch acts as the student model in our KD framework. These networks, serving as teacher and student, respectively, do not share weights.

\subsection{Cross-Modality Alignment}
\label{subsec: CMA}
The objective of CMA is to enhance the density of radar BEV features, thereby facilitating the transfer of knowledge from LiDAR features to radar features more effectively.
We note that the average number of non-empty pillars in radar features $F^{(l)}_{rdr}$ is only approximately 10\% of the average number found in LiDAR features $F^{(l)}_{ldr}$. This considerable difference in the number of non-empty pillars poses challenges in aligning features between the two modalities. Specifically, transferring information from non-empty pillars in LiDAR data to corresponding empty pillars in radar data proves to be impractical. To tackle this challenge, CMA is employed to densify radar features. Our empirical study demonstrates that this densification process significantly contributes to the successful knowledge distillation from LiDAR to radar.

\begin{figure}[t]
\centering  
\includegraphics[width=7.5cm]{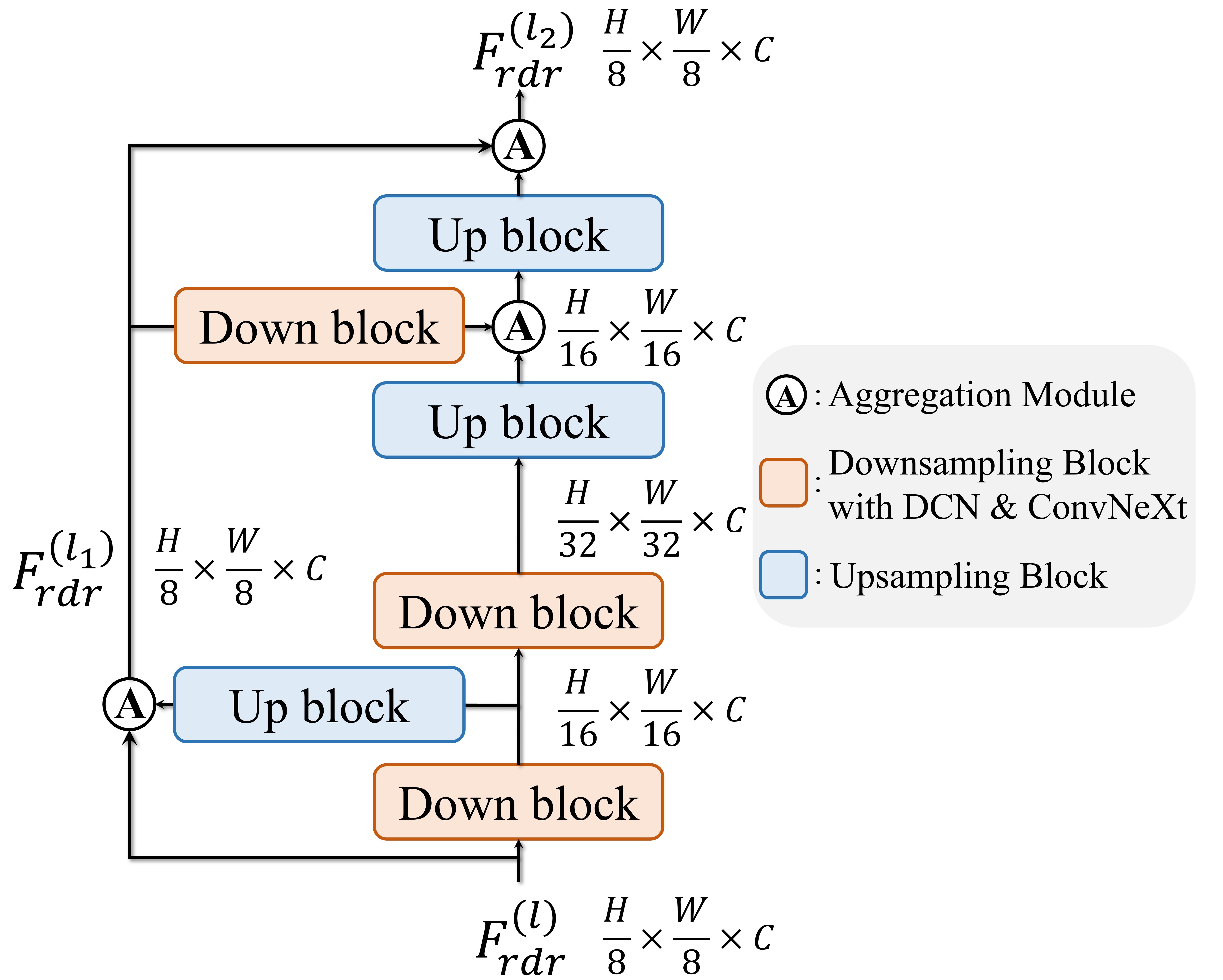}
\caption{\textbf{Detailed structure of the proposed CMA module.}}
\label{CMA_architecture}  
\end{figure}

Fig. \ref{CMA_architecture} illustrates the architecture of CMA. CMA comprises the  {\it Down Block}, {\it Up Block}, and {\it Aggregation Module}.
{\it Down Block} conducts down-sampling via deformable convolution \cite{deformconv}, followed by the ConvNeXt V2 blocks \cite{convnextv2}. The {\it Up Block} conducts up-sampling operations using a 2D transposed convolution. The {\it Aggregation Module} concatenates two input features and applies a 1$\times$1 convolution layer.
After applying the {\it Down Block} and {\it Up Block} operations twice each, in addition to incorporating a side pathway through the {\it Aggregation Module}, CMA generates two low-level BEV features, $F^{(l_1)}_{rdr}$ and $F^{(l_2)}_{rdr}$, at a $1/8$ resolution. As a result of CMA's dilation operation, these low-level features exhibit increased density compared to the input radar features. (Refer to Fig. \ref{architecture} for a visual comparison of the features before and after CMA.) These intermediate features are subsequently utilized for knowledge distillation in the following AFD stage. For a detailed structure of the CMA, refer to the Supplementary Materials.

\subsection{Activation-based Feature Distillation}
\label{subsec: AFD}

AFD conducts Activation-aware Feature Matching on the low-level features obtained from both radar and LiDAR branches.
AFD performs knowledge distillation from the LiDAR BEV features $F^{(l)}_{ldr}$ to the first radar BEV features $F^{(l_1)}_{rdr}$. Simultaneously, it also performs knowledge distillation from $F^{(l)}_{ldr}$ to the second radar BEV features $F^{(l_2)}_{rdr}$.

AFD conducts selective knowledge distillation based on regions using active masks.
We determine the active masks \(M^{(l_1)}_{rdr}\), \(M^{(l_2)}_{rdr}\), and \(M^{(l)}_{ldr}\) using the corresponding densified radar features \(F^{(l_1)}_{rdr}\), \(F^{(l_2)}_{rdr}\), and the low-level LiDAR features \(F^{(l)}_{ldr}\), respectively, as
 \begin{align}
 & F^{(l')}_{mod,i,j}=\sum^{C}_{c=1}F^{(l')}_{mod,c,i,j}, \label{eq:fmod1}\\ \label{eq:fmod2}
 & M^{(l')}_{mod,i,j}=
 \begin{cases}
 1, &{\rm if} \, F^{(l')}_{mod,i,j} > 0, \\
 0, &{\rm otherwise},
 \end{cases}
\end{align}
where $F \in \mathbb{R}^{C \times \frac{H}{8} \times \frac{W}{8}}$ represents BEV features, $i, j, $ and $c$ correspond to height, width, and channel indices, respectively, $mod\in\{ldr, rdr\}$, and $l' \in \{l_1, l_2, l\}$.

Using the active masks obtained from (\ref{eq:fmod1}) and (\ref{eq:fmod2}), we can identify the active regions (AR) where both radar and LiDAR features are active, and the inactive regions (IR) where radar features are active and LiDAR features are inactive, i.e., 
\begin{align}   
 AR^{(l_n)} &=\,(M^{(l_n)}_{rdr}=1)\,\&\,(M^{(l)}_{ldr}=1), \\
 IR^{(l_n)} &=\,(M^{(l_n)}_{rdr}=1)\,\&\,(M^{(l)}_{ldr}=0), 
\end{align}
where $n\in [1,2]$.
AFD minimizes the difference between radar and LiDAR features in both the AR and IR. AFD is specifically trained to transform radar features to resemble LiDAR features within AR, while simultaneously learning to suppress radar features within IR.

Due to the sparsity of LiDAR data, the AR area is typically much smaller than the IR area. To prevent the model from overly focusing on knowledge distillation within IR, we scale the loss terms associated with AR and IR based on their relative area ratios.
We devise adaptive loss weights $W^{(l_n)}_{sep}$ to balance the loss terms within AR and IR according to their respective pixel counts, i.e.,  
\begin{align}
 W^{(l_n)}_{sep,i,j} &=
 \begin{cases}
 \alpha, & if \, (i,j) \in AR^{(l_n)}, \\
 \rho^{(l_n)}\times \beta, & if \, (i,j) \in IR^{(l_n)}, \\
 0, & \, otherwise,
 \end{cases} 
\label{eq:decouple}
\end{align}
where $\rho^{(l_n)}=\frac{N_{AR^{(l_n)}}}{N_{IR^{(l_n)}}}$ represents the relative importance of the $IR^{(l_n)}$ over $AR^{(l_n)}$, $\alpha$ and $\beta$ are the intrinsic balancing parameters, and $N_{AR^{(l_n)}}$ and $N_{IR^{(l_n)}}$ are the number of pixels in $AR^{(l_n)}$ and $IR^{(l_n)}$, respectively. The weight associated with IR is determined proportionally by $\rho$, representing the ratio of pixel count for AR to that for IR.

The distillation loss $L^{(n)}_{low}$ for the $n$-th activation features is given by
\begin{align}   
 L^{(n)}_{low} &= \sum^{C}_{c=1}\sum^{H}_{i=1}\sum^{W}_{j=1} W^{(l_n)}_{sep,i,j}(F^{(l)}_{ldr,c,i,j}-F^{(l_n)}_{rdr,c,i,j})^2.
\end{align}
The final AFD loss $L_{AFD}$ is obtained by averaging the distillation loss functions associated with two densified radar features \(F^{(l_1)}_{rdr}\) and \(F^{(l_2)}_{rdr}\)
\begin{align}   
 L_{AFD} &= \frac{1}{2}\sum^{2}_{n=1}L^{(n)}_{low}.
\end{align}

\subsection{Proposal-based Feature Distillation}
\label{subsec: PFD}
PFD employs Proposal-level Feature Matching to transfer knowledge from the high-level features of LiDAR to those of radar. This guidance helps the radar network generate object features that closely mimic the high-level features of LiDAR, within object proposals.

DenseEnc is applied to produce two high-level features $F^{(h_1)}_{ldr}$ and $F^{(h_2)}_{ldr}$ from the low-level features $F^{(l)}_{ldr}$ in the LiDAR branch. Similarly, DenseEnc also produces two high-level features $F^{(h_1)}_{rdr}$ and $F^{(h_2)}_{rdr}$ from $F^{(l_2)}_{rdr}$ in the radar branch. DenseEnc generates the first high-level features $F^{(h_1)}_{mod}$ using the Conv2D-BN-ReLU block followed by a 2D convolution block and a 2D transposed convolution. Here, the Conv2D-BN-ReLU block refers to a sequence of operations comprising 2D Convolution (Conv2D), Batch Normalization (BN), and Rectified Linear Unit (ReLU), while the 2D convolution block consists of six layers of Conv2D-BN-ReLU. 
For LiDAR branch, DenseEnc produces the second high-level LiDAR features $F^{(h_2)}_{ldr}$ by concatenate \(F_{ldr}^{(h_1)}\) with \(F_{ldr}^{(l)}\) and applying the 2D Convolution block again. DenseEnc also generates the second high-level radar features $F^{(h_2)}_{rdr}$ by concatenate \(F_{rdr}^{(h_1)}\) with \(F_{rdr}^{(l_2)}\) and applying the 2D Convolution block in radar branch.
Finally, the CenterHead is applied to both $F^{(h_2)}_{ldr}$ and $F^{(h_2)}_{rdr}$ to produce the classification heatmaps $H^{cls}_{ldr}$ and $H^{cls}_{rdr}$, respectively.

PFD conducts knowledge distillation solely within the key target regions of high-level BEV features.
PFD identifies key target regions  based on the predicted radar heatmap $H^{cls}_{rdr}$ and the ground truth heatmap $H^{cls}_{GT}$. The ground truth heatmap is generated by projecting the 3D centers of ground truth boxes onto the BEV space and applying a Gaussian kernel to represent object locations \cite{centernet}. The identified key target regions are labeled as true positives (TP), false positives (FP), or false negatives (FN) according to 
\begin{align}   
 TP &=\,(H^{cls}_{GT}>\sigma)\,\&\,(H^{cls}_{rdr}>\sigma), \\
 FP &=\,(H^{cls}_{GT}<\sigma)\,\&\,(H^{cls}_{rdr}>\sigma), \\
 FN &=\,(H^{cls}_{GT}>\sigma)\,\&\,(H^{cls}_{rdr}<\sigma),
\end{align}
where $\sigma$ represents the threshold parameter set to 0.1 in our setup.
TP and FN regions encompass areas containing real objects, while FP regions correspond to mis-detected regions. Consequently, PFD concentrates on feature alignment in both TP and FN regions while suppressing radar features in FP regions. Due to the inherent imbalance in the proportions of these three regions, distinct weights are applied to the distillation loss terms associated with TP, FN, and FP regions.
The proposal-dependent loss weights $W_{proposal}$ are determined depending on the area of each region
\begin{align}   
 W_{proposal,i,j} &=
 \begin{cases}
 \frac{\lambda_1}{N_{TP}+N_{FN}}, & if \, (i,j) \in (TP \cup FN), \\
 \frac{\lambda_2}{N_{FP}}, & if \, (i,j) \in FP, \\
 0, & \, otherwise,
 \end{cases}
\label{eq:pfd_weight}
\end{align}
where $\lambda_1$ and $\lambda_2$ represent the balancing parameters for the respective regions, and $N_{TP}$, $N_{FN}$ and $N_{FP}$ denote the number of pixels within TP, FN, and FP regions, respectively.

Finally, the distillation loss defined for the $m$th high-level features is weighted by the loss weights $W_{proposal}$ as
 \begin{align}
L^{(m)}_{high}&=\sum^C_{c=1}\sum^H_{i=1}\sum^W_{j=1}W_{proposal,i,j}\left|S^{(h_m)}_{ldr,c,i,j}-S^{(h_m)}_{rdr,c,i,j}\right|, 
    \end{align}
where
\begin{align}
S^{(h_m)}_{mod,c,i,j}=\frac{\text{exp}(F^{(h_m)}_{mod,c,i,j})}{\sum^{C}_{k=1}\text{exp}(F^{(h_m)}_{mod,k,i,j})},
\end{align}
where $mod \in \{rdr, ldr \}$. 
Both the radar and LiDAR features $F^{(h_m)}_{rdr}$ and $F^{(h_m)}_{ldr}$ go through normalization in the channel dimension, resulting in $S^{(h_m)}_{rdr}$ and $S^{(h_m)}_{ldr}$, respectively. This normalization step aims to align the magnitude scale of the high-level features between the radar and LiDAR branches. 
The final PFD loss $ L_{PFD}$ is computed by averaging the distillation loss 
 over $m=1,2$, i.e., 
\begin{align}   
 L_{PFD}=\frac{1}{2}\sum^2_{m=1}L^{(m)}_{high}.
\end{align}

\subsection{Loss Function}
\label{subsec: Overall Loss Function}
The total loss function used to train RadarDistill is obtained from 
\begin{align}
 L_{total}=L_{det} + \gamma L_{AFD} + \delta L_{PFD},
\label{eq:total_loss}
\end{align}
where $\gamma$ and $\delta$ are the regularization parameters for weighting the loss terms $L_{AFD}$ and $L_{PFD}$, respectively.

\section{Experiments}
\subsection{Experimental Setup}
\noindent{\bf Datasets and metrics.} We conduct the experiments on the nuScenes \cite{nuscenes} dataset. This dataset contains 700, 150, 150 driving scenes for training, validation, and testing, respectively.
The nuScenes dataset gathers radar data using three radar sensors at the vehicle’s front and corners, and two at the rear, providing 360-degree coverage. These sensors collectively capture data at a frequency of 13Hz, ensuring broad coverage up to 250 meters and offering velocity measurement accuracy within $\pm 0.1 km/h$. 
Our evaluation utilizes the official metrics from nuScenes, which include mean average precision (mAP) and the nuScenes detection score (NDS).

\begin{table*} [ht!] 
    \centering
    \begin{tabular}{c|cc|cc|ccccc}
        \noalign{\smallskip}\noalign{\smallskip}\toprule[1.0pt]
        Method & Input & KD & mAP$\uparrow$ & NDS$\uparrow$ & mATE$\downarrow$ & mASE$\downarrow$ & mAOE$\downarrow$ & mAVE$\downarrow$ & mAAE$\downarrow$ \\\hline
        Radar-PointGNN \cite{radar-pointgnn}   & R & - & 0.5 & 3.4 & 1.024 & 0.859 & 0.897 & 1.020 & 0.931 \\
        KPConvPillars \cite{kpconvpillar}       & R & - & 4.9 & 13.9 & 0.823 & 0.428 & 0.607 & 2.081 & 1.000 \\
        PillarNet* \cite{pillarnet} (Our baseline)  & R & - & 8.6 & 34.7 & 0.532 & 0.283 & 0.615 & 0.438 & 0.092 \\\Xhline{0.9pt}
        \bf RadarDistill & R & \checkmark & \bf 20.5 & \bf 43.7 & \bf 0.461 &  \bf 0.263 & \bf 0.525 & \bf 0.336 & \bf 0.072 \\\bottomrule[1.0pt]
    \end{tabular}
    \caption{\textbf{Performance evaluation on {\it nuScenes testset}.} The best performed metrics in the radar-only model are marked in bold. ``*'' denotes models we have reproduced without applying test time augmentation. Our model achieves the state-of-the-art performance in all metrics.} 
    \label{nuscenes test result}
\end{table*}

\begin{table*} [ht] 
    \centering
    \scalebox{1.0}
    {
    \begin{tabular}{c|cc|c@{\hspace{2mm}}c@{\hspace{2mm}}c@{\hspace{2mm}}c@{\hspace{2mm}}c@{\hspace{2mm}}c@{\hspace{2mm}}c@{\hspace{2mm}}c@{\hspace{2mm}}c@{\hspace{2mm}}c}
        \noalign{\smallskip}\noalign{\smallskip}\toprule[1.0pt]
        Method & Input & KD & Car & Truck & bus & Trailer & C.V & Ped. & Motor. & Bicycle & T.C & Barrier \\\hline
        CenterFusion \cite{centerfusion}   & C, R & - & 50.9 & 25.8 & 23.4 & 23.5 & 7.7 & 37.0 & 31.4 & 20.1 & 57.5 & 48.4 \\
        FCOS3D \cite{fcos3d}       & C & - & 52.4 & \textbf{27.0} & \textbf{27.7} & 25.5 & \textbf{11.7} & \textbf{39.7} & \textbf{34.5} & \textbf{29.8} & 55.7 & \textbf{53.8} \\
        MonoDIS \cite{monodis}       & C & - & 47.8 & 22.0 & 18.8 & 17.6 & 7.4 & 37.0 & 29.0 & 24.5 & 48.7 & 51.1 \\
        CenterNet \cite{centernet}       & C & - & 53.6 & \textbf{27.0} & 24.8 & 25.1 & 8.6 & 37.5 & 29.1 & 20.7 & \textbf{58.3} & 53.3 \\
        PillarNet* \cite{pillarnet}  (Our baseline)  & R & - & 41.8 & 11.6 & 8.4 & 6.5 & 0.0 & 7.4 & 1.0 & 0.0 & 1.1 & 8.1 \\\Xhline{0.9pt}
        \bf RadarDistill & R & \checkmark & \textbf{54.0} & 15.3 & 11.3 & \textbf{29.5} & 5.5 & 9.2 & 15.3 & 0.9 & 21.7 & 42.3 \\\bottomrule[1.0pt]
    \end{tabular}
    }
    \caption{\textbf{Performance evaluation per class on {\it nuScenes testset}.} The best performed metrics are marked in bold. `C.V', `Ped.', `Motor.', and `T.C.' represent construction vehicle, pedestrian, motorcycle, and traffic cone, respectively.} 
    \label{nuscenes per-class result}
\end{table*}

\noindent{\bf Implementation details.} As our baseline model, we employed PillarNet-18, which corresponds to PillarNet \cite{pillarnet} with a ResNet-18 backbone. We utilized the Adam optimizer with a learning rate of 0.001, implementing a one-cycle learning rate policy. The weight decay was set to 0.01 with the momentum scheduled from 0.85 to 0.95. We trained the PillarNet-18 model for 20 epochs with a batch size of 16. We used the {\it Class-Balanced Grouping and Sampling} (CBGS) strategy \cite{cbgs} to mitigate class imbalance issue. Data augmentation techniques were applied, including random scene flipping along the $X$ and $Y$ axes, random rotation, scaling, translation, and ground-truth box sampling. Our detection range was set to $[-54m, 54m]$ for both the $X$ and $Y$ axes, and $[-5m, 3m]$ for the $Z$ axis. We used pillars of dimensions $(0.075m, 0.075m)$.

Our proposed RadarDistill model was trained for 40 epochs. The rest of training setup follows the same training setup as the baseline model. We initialized the radar backbone network using the pre-trained LiDAR backbone network, following the {\it Inheriting Strategy} presented in \cite{kang2021instance}. The hyperparameters $\alpha$ and $\beta$ in (\ref{eq:decouple}) were set to $3 \times 10^{-4}$ and $5 \times 10^{-5}$, respectively. Additionally, the hyperparameters $\lambda_1$ and $\lambda_2$ in (\ref{eq:pfd_weight}) were set to $5$ and $1$, respectively. The parameters $\gamma$ and $\delta$ in (\ref{eq:total_loss}) were set to $5$ and $25$, respectively. The entire training  was conducted on 4 NVIDIA RTX 3090 GPUs.

\subsection{Performance Comparison}
Table \ref{nuscenes test result} presents the performance of RadarDistill on {\it nuScenes testset}.
Our RadarDistill achieves significant performance improvements compared to other radar-based object detectors. Specifically, RadarDistill achieves a mAP of 20.5\% and a NDS  of 43.7\%, outperforming the previous state-of-the-art (SOTA) method, KPConvPillars \cite{kpconvpillar}, by a considerable margin of +15.6\% in mAP and +29.8\% in NDS. 
Table \ref{nuscenes per-class result} displays the AP performance of RadarDistill for each class. We compare our RadarDistill with other camera-only and camera-radar fusion-based object detectors. RadarDistill exhibits superior performance compared to the baseline, PillarNet-18 \cite{pillarnet} across all classes. Notably, it achieves performance gains of +12.2\% and +23\% for the Car and Trailer classes, respectively. Furthermore, RadarDistill outperforms other camera-only and sensor fusion models in the Car class, which is surprising given the limited resolution of radar data compared to camera data.

\subsection{Ablation Studies}
We conducted ablation studies on the nuScenes validation set to assess the impact of each proposed idea. Our model was trained using $1/7$ of the training set and evaluated on the entire validation set.

\noindent{\bf Component Analysis.} Table \ref{tab:module_ablation} illustrates the performance enhancements achieved by each component of our proposed model. Integrating CMA into the PillarNet-18 baseline results in a 2\% improvement in NDS performance. With both CMA and AFD enabled, the NDS performance further improves by 4.4\%. Activating PFD yields a 1.0\% additional increase in NDS performance. We also tried disabling CMA while other components are enabled in RadarDistill. In this case, the NDS performance drops by 4\%, highlighting the critical role of CMA in our RadarDistill model.

\noindent{\bf Proposed Distillation Method.}
We compare  the proposed Distillation method with other well known KD methods.
We consider the following methods
\begin{itemize}
\item {\it Baseline}*: This baseline is constructed by integrating CMA into the PillarNet-18 baseline.
\item {\it Baseline}**: This baseline is constructed by applying CMA and AFD to the PillarNet-18 baseline.
\item {\it Complete} \cite{fitnets}: This method applies the fixed weight across entire areas of the low-level BEV features.
\item {\it Gaussian} \cite{TADF}: This approach generates a 2D Gaussian mask based on the centers of GT boxes and assigns higher weighted distillation loss to areas closer to the foreground.
\item {\it FG/BG} \cite{defeat}: This strategy separates the foreground and background using GT boxes, applying different weighted distillation losses for each.
\end{itemize}
Table \ref{tab:region_afd_ablation} evaluates the performance of AFD in comparison with other KD methods. 
We applied AFD and these methods to the Baseline*, where  the low-level BEV features are produced by CMA. 
We observe that the  proposed {\it AFD} achieves significantly better performance than other KD methods. 
Particularly, it achieves the performance gains of 3.7\% in AP in car category, 1.3\% in mAP, and 2.0\% in NDS over the {\it FG/BG} method.

Table \ref{tab:region_dfd_ablation} compares PFD with other KD methods when applied to high-level BEV features. We applied these methods to Baseline** where the high-level features are produced by  CMA and AFD. We also confirm that the proposed PFD achieves higher detection performance gains than other KD methods. Specifically, it yields the performance improvements of 0.5\% in AP for the car category, 0.2\% in mAP, and 1.0\% in NDS over the {\it FG/BG} method.

\begin{table}[t]
  \centering
  \fontsize{10pt}{12pt}\selectfont
  \scalebox{1.0}
  {
  \begin{tabular}{ccc|cc}
    \toprule[1.0pt]
         \textit{CMA} & \textit{AFD} & \textit{PFD} & mAP$\uparrow$ & NDS$\uparrow$ \\
        \midrule
          &  &   & 5.4 & 27.3 \\
         \checkmark &  &  & 6.4 & 29.3 \\
         \checkmark & \checkmark &   & 10.9 & 33.7 \\
         \checkmark & \checkmark & \checkmark & \textbf{11.2} & \textbf{34.7} \\
          & \checkmark & \checkmark  & 7.0 & 30.7 \\
       \bottomrule[1.0pt]
  \end{tabular}
  }
  \caption{\textbf{Ablation study for evaluating the contribution from each component of Radardistill on {\it nuScenes validation set}.} 
  }
  \label{tab:module_ablation}
\end{table}

\begin{table}[t]
  \centering
  \fontsize{10pt}{12pt}\selectfont
  \scalebox{0.98}
    {
    \begin{tabular}{c|ccc}
        \toprule[1.0pt]
        \multirow{2}{*}{Region} & \multicolumn{3}{c}{\textit{AFD}} \\ \cline{2-4}
        & Car(AP)$\uparrow$ & mAP$\uparrow$ & NDS$\uparrow$ \\
        \midrule
        Baseline* & 34.8 & 6.4 & 29.3 \\
        Complete \cite{fitnets} & 39.0 & 8.6 & 32.2 \\
        Gaussian \cite{TADF} & 40.4 & 8.8 & 29.5 \\
        FG/BG \cite{defeat} & 42.1 & 9.6 & 31.7 \\
        Our AFD & \textbf{45.8} & \textbf{10.9} & \textbf{33.7} \\
        \bottomrule[1.0pt]
    \end{tabular}
    }
    \captionof{table}{\textbf{Comparison of AFD with different KD methods evaluated on {\it nuScenes validation set}.}}
    \label{tab:region_afd_ablation}
\end{table}



\noindent{\bf Impact of Scale Normalization in PFD.} Next, we investigate the impact of scale normalization in PFD.  Table \ref{tab:scale_alignment_pfd} presents a comparison of performances with and without scale normalization.
We observe that our scale normalization results in a significant improvement of 6.2\% in AP (CAR category), 2.7\% in mAP, and 2.9\% in NDS. This underscores the substantial difference in magnitude scale between radar and LiDAR high-level features, and demonstrates that narrowing this gap significantly boosts the effect of knowledge distillation.

\noindent{\bf Comparison when applied to Radar-Camera Fusion.}
Next, we investigate whether the improved features generated by RadarDistill can also lead to performance enhancements in radar-camera fusion methods. Table \ref{tab:rcfusion} presents the detection performance when the proposed RadarDistill is intergrated into the  radar-camera fusion method, BEVFusion \cite{bevfusion}. Because BEVFusion was originally designed for LiDAR-camera fusion, we adapted its design for radar-camera fusion following  \cite{crn}. Then, RadarDistill was simply incorporated to the radar encoding network in BEVFusion and the entire model was trained end to end.
We note that RadarDistill yields performance improvements of 1.8\% in Car AP, 1.3\% in mAP, and 1.1\% in NDS over the baseline BEVFusion model. We believe that further enhancements in performance could be achieved by applying more sophisticated fusion strategies, a direction we leave for future research.

\begin{table}[t]
  \centering
  \fontsize{10pt}{12pt}\selectfont
  \scalebox{0.96}
    {
    \begin{tabular}{c|ccc}
        \toprule[1.0pt]
        \multirow{2}{*}{Region} & \multicolumn{3}{c}{\textit{PFD}} \\ \cline{2-4}
        & Car(AP)$\uparrow$ & mAP$\uparrow$ & NDS$\uparrow$ \\
        \midrule
        Baseline** & 45.8 & 10.9 & 33.7 \\
        Complete \cite{fitnets} & 45.2 & 10.8 & 33.8 \\
        Gaussian \cite{TADF} & 45.9 & 10.8 & 33.8 \\
        FG/BG \cite{defeat} & 45.6 & 11.0 & 33.7  \\
        Our PFD & \textbf{46.1} & \textbf{11.2} & \textbf{34.7} \\
        \bottomrule[1.0pt]
    \end{tabular}
    }
    \captionof{table}{\textbf{Comparison of PFD with different KD methods evaluated on {\it nuScenes validation set}.}}
    \label{tab:region_dfd_ablation}
\end{table}

\begin{table}[t]
  \centering
  \fontsize{10pt}{12pt}\selectfont
  \scalebox{0.9}
  {
  \begin{tabular}{c|ccc}
    \toprule[1.0pt]
        Method & Car(AP)$\uparrow$ & mAP$\uparrow$ & NDS$\uparrow$ \\
        \midrule 
        PFD w/o norm  & 39.9 & 8.5 & 31.8 \\
        PFD & \textbf{46.1} & \textbf{11.2} & \textbf{34.7} \\
        \bottomrule[1.0pt]
  \end{tabular}
  }
  \caption{\textbf{Ablation study for evaluating the impact of scale normalization used in PFD on {\it nuScenes validation set}.}  {\it PFD w/o norm}:  PFD without scale normalization. {\it PFD}: PFD with scale normalization.}
  \label{tab:scale_alignment_pfd}
\end{table}

\begin{table}[t]
  \centering
  \fontsize{10pt}{12pt}\selectfont
  \scalebox{0.85}
  {
  \begin{tabular}{c|cc|ccc}
    \toprule[1.0pt]
        Method & Input & KD & Car(AP)$\uparrow$ & mAP$\uparrow$ & NDS$\uparrow$ \\
        \midrule
        BEVFusion* \cite{bevfusion} & C,R & & 65.9 & 38.3 & 45.3 \\
        RadarDistill-CR & C,R & \checkmark & \textbf{67.7} & \textbf{39.6} & \textbf{46.4} \\
        \bottomrule[1.0pt]
  \end{tabular}
  }
  \caption{\textbf{Ablation study for evaluating the impact of RadarDistill on radar-camera fusion models.} ``*'' denotes models we have reproduced without applying test time augmentation. }
  \label{tab:rcfusion}
\end{table}

\section{Conclusions}
In this study, we introduced RadarDistill, a novel radar-based 3D object detection method aimed at enhancing radar data features by leveraging information from LiDAR data through knowledge distillation. We proposed three effective techniques—CMA, AFD, and PFD—to make the challenging task of transferring knowledge from LiDAR to radar successful. Our approach emphasizes guiding the radar encoding network to generate features that closely resemble the semantically rich features of LiDAR. 
CMA densifies radar features, aiding the radar encoding network in learning the complex distribution of LiDAR features better. AFD and PFD target the reduction of discrepancies between radar and LiDAR features in both low-level and high-level BEV features.
Our evaluation demonstrated that RadarDistill can yield significant performance improvements over both radar-based and radar-camera fusion-based baselines, establishing state-of-the-art performance in radar-only object detection.

\section{Acknowledgement}
This work was partly supported by Institute of Information \& communications Technology Planning \& Evaluation (IITP) grant funded by the Korea government (MSIT) [No.2021-0-01343-004, Artificial Intelligence Graduate School Program (Seoul National University)] and the National Research Foundation of Korea (NRF) grant funded by the Korea government (MSIT) [No.2020R1A2C2012146].

{\small
\bibliographystyle{ieee_fullname}
\bibliography{main}
}
\end{document}